\documentclass[letterpaper, 10 pt, conference]{ieeeconf}  
\IEEEoverridecommandlockouts
\usepackage{cite}
\usepackage{amsmath,amssymb,amsfonts}
\usepackage{algorithmic}
\usepackage{graphicx}
\usepackage[hyphens,spaces,obeyspaces]{url}
\usepackage{hyperref}
\usepackage{todonotes}
\usepackage{textcomp}
\usepackage{xcolor}
\usepackage{bm}
\usepackage{threeparttable}
\usepackage{siunitx}
\usepackage{multirow}
\usepackage{subfig}
\usepackage{comment}
\usepackage{tikz}
\usepackage{url}

\usepackage{caption}
\usepackage{hyperref}
\usepackage{lipsum}
\usepackage{gensymb}
\usetikzlibrary{shapes}

\usepackage[skip=8pt,font=small]{caption}

\def\BibTeX{{\rm B\kern-.05em{\sc i\kern-.025em b}\kern-.08em
    T\kern-.1667em\lower.7ex\hbox{E}\kern-.125emX}}

\hypersetup{
	colorlinks=true,
	linkcolor=black,
	filecolor=magenta,      
	urlcolor=black,
}

\begin{document}

\title{\LARGE \bf Analysis of Mutual and Referential Human and Robot Gazes in a Collaborative Word Association Game




}

\author{Jens V. Rüppel$^{1}$,
       Tim Schreiter$^{1,3}$,
       Andrey Rudenko$^1$, 
       and~Achim J.~Lilienthal$^{1,2,3}$
\thanks{$^{1}$Chair of Perception for Intelligent Systems, Munich Institute of Robotics and Machine Intelligence (MIRMI), Technical University of Munich (TUM), Germany {\tt\small \{jens.v.rueppel, tim.schreiter, andrey.rudenko, achim.j.lilienthal\}@tum.de}}
\thanks{$^{2}$Centre for Applied Autonomous Sensor Systems (AASS), Örebro University, Sweden}
\thanks{$^{3}$Robotics Insitute Germany (RIG), Germany}}

\maketitle

\begin{abstract}

Robot gaze is a major component of human-robot dialogue coordination. Most studies of gaze in human-robot dialogue focus on face-to-face social conversations, but little is known about gaze in demanding task-focused interactions. In this paper, we investigate how the gaze of a robot game partner affects human visual attention and if humans tend to direct confirmation-seeking gazes towards the robot. In our study, we let participants play a collaborative word association game with a NAO robot acting as an embodied, LLM-driven conversational partner. Our experiments are conducted under two conditions, which implement mutual and referential gazes of the robot respectively. We record participants' gaze using eye tracking glasses and analyze the interactions using gaze coordinates, speech segments, key events and areas of interests. We find that robot gaze orientation does not affect the time to first fixation on words the robot proposed. We also find that participants gaze more often at the robot when their dialogue line contains confirmation requests, compared to when it does not. Our results indicate (likely also due to the cognitively demanding nature of the game) that the verbal aspect of this task overshadows the effects of referential robot gaze.
These findings offer valuable insights for designing and validating robot gaze and turn-taking behavior in collaborative tasks which require coordination and efficient communication. 

\end{abstract}

\section{Introduction}\label{sec:introduction}


Dialogue between humans and social robots is inherently multi-modal. One part is the verbal interaction, which involves speech, turn-taking and pacing. The other part is non-verbal behavior such as gestures and gaze. The latter in particular serves an important role throughout a conversation. Admoni et al. \cite{admoni2017social} describe two important forms of gaze, mutual and referential, serving different functions. \textit{Mutual gaze}, which is established between conversation partners, supports the social exchange through seeking feedback and managing turn-taking \cite{kendon1967some, ho2015speaking}. Mutual gaze is used to assure that joint attention is established and a joint understanding of the task is reached \cite{moore2014joint}. However, mutual gaze can also potentially slow down decision-making and response times \cite{belkaid2021mutual}. \textit{Referential gaze} directs attention to the relevant task elements \cite{moore2014joint} and induces reflexive gaze shifts of the interaction partner towards the referenced object \cite{ristic2007attentional}. In contrast to both, gaze aversion (avoiding either of the two) can also occur in a conversation due to elevated cognitive efforts, e.g. when answering difficult questions \cite{glenberg1998averting}.



\begin{figure}[t!]
    \centering
    \includegraphics[width=\columnwidth]{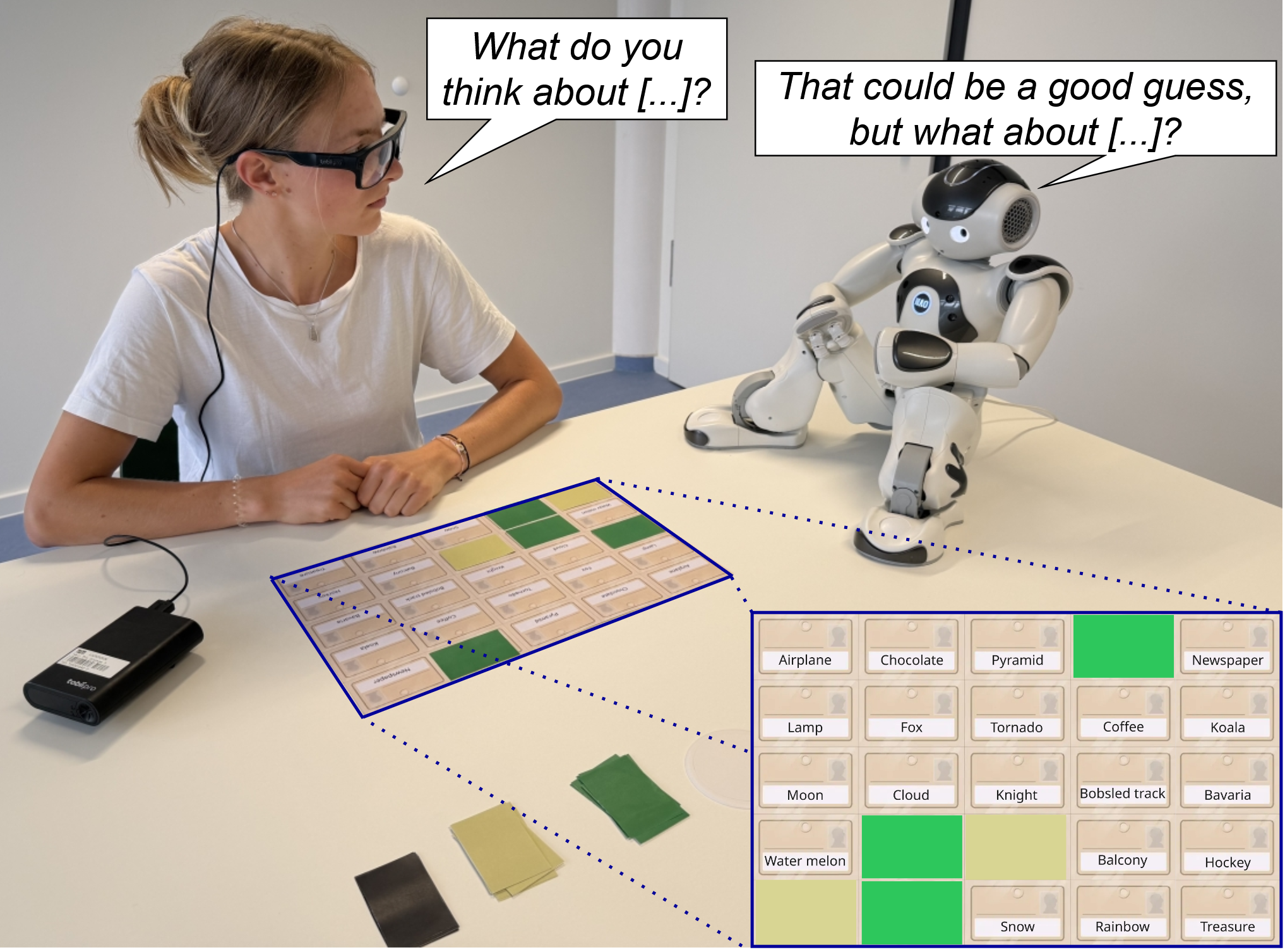}
    \caption{A participant is playing the collaborative word association game with the robot, while sharing their attention between the partner and the board. The robot operates in two conditions: (1) referential gaze, e.g., looking at the board, and (2) mutual gaze, e.g., looking at the participant. The gaze of the participants, as well as the dialogue lines of the participant and the robot, is recorded with eye tracking glasses.}
    \label{fig:frontal}
\end{figure}


Methods for interpreting human gazes and producing human-like robot gaze behavior have an effect on the communication smoothness, legibility and comfort of the human participant. These non-verbal aspects of the dialogue can raise interest in engaging with the robot \cite{bruce2002role}, reduce reaction times \cite{schreiter2023advantages}, improve handover performance \cite{fischer2015effects, lavit2024gaze} and establish joint attention \cite{yu2012adaptive}, especially when people believe that the robot can see the target \cite{morillo2023can}.
Especially in collaborative tasks, team performance can depend on the communication quality. When working with robots, human attention tends to be divided between task-relevant objects and the robot's embodiment and they spend more time looking at a robot than at a human in a corresponding human-human task \cite{yu2012adaptive}. However, a robot gaze does not reliably trigger cueing effects like human gaze would \cite{onnasch2022humans, admoni2011robot} and gaze cueing for the robot only emerges after establishing eye contact \cite{kompatsiari2018role}.

\begin{figure*}[t!]
    \centering
    \vspace{0.2cm}
    \includegraphics[width=\columnwidth*2]{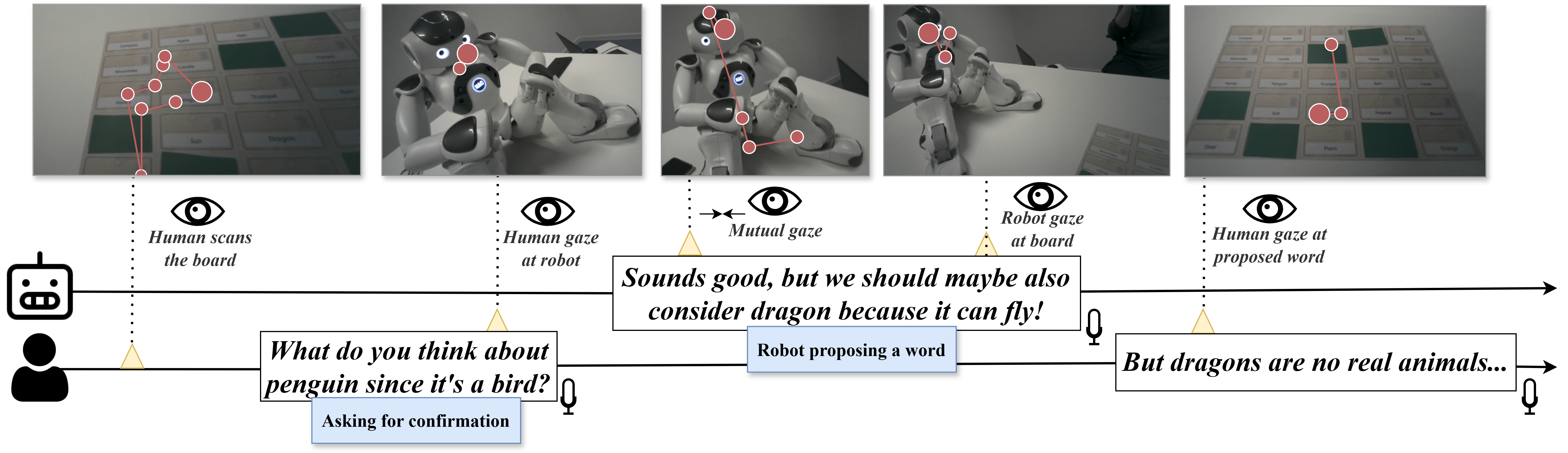}
    \caption{Timeline of an example discussion about the clue ``Birds, 2'', including speech and gaze events}
    \label{fig:timeline_interaction}
\end{figure*}

Games offer a controlled and socially-rich environment to study the gaze behavior of humans and the effect of robot gaze \cite{rato2023robots}. For example, in a competitive conversation-free game, mutual robot gaze tends to delay human decision-making and increase reaction times \cite{belkaid2021mutual}. The survey of Rato et al. \cite{rato2023robots} of robots in games shows that robot game partners in the literature rarely perceive or respond to social cues such as speech or gaze. 
Therefore, collaborative dialogue-driven scenarios, where high task demand can reduce mutual gaze in the first place, remain largely underexplored.

In this paper, we conduct a study of mutual and referential gazes of robots and people in a cognitively-demanding collaborative game. We propose a game which requires verbal coordination between the robot and the participant to make a joint decision about the cards placed on the game board. To analyze the interaction, we record first-person videos and gaze fixation points of the participant during the game, and align them with the dialogue lines, labeled key events and annotated areas of interest on the game board and in the environment.

Our study is motivated by two research questions (RQ):
\begin{itemize} 
\item \textbf{RQ 1:} Does confirmation-seeking behavior of the participant trigger increased gaze towards the robot?
\item \textbf{RQ 2:} How does the robot's gaze behavior affect the time to first fixation towards the target?
\end{itemize}





These two RQs have implications in other task-focused verbal interactions between robots and people, for instance in collaborative assembly or in assistive healthcare. In order to investigate those RQs, we implement two conditions, in which the gaze behavior of the robot is manipulated: one condition with more mutual than referential gaze, the other condition vice versa. We compared the gaze behavior of the participant between those conditions.

\section{Methods}\label{sec:methods}

\subsection{Game design}

Our game is a word association task inspired by the board game Codenames\footnote{https://www.czechgames.com/games/codenames}. In our implementation, one participant plays in cooperation with the robot, and the experimenter acts as the ``game master". The game is played in English. A board with a grid of words is placed in front of the participant and the robot. Each word is either a hit, a miss or a ``game over'' event which immediately ends the game. In each turn, the game master first proposes a clue, e.g. ``Colorful, 3'', where the number indicates the amount of words the clue should be associated with. Then the participant and the robot discuss possible words associated with the clue word. After the discussion is over, the participant points to the respective word (minimum 1, maximum the amount of cues) and a card with either a green (hit), a yellow (miss) or a black (game over) color is placed on top of the word at the board. We show an example interaction in Figure \ref{fig:frontal}.

\begin{figure}[t!]
   \centering
   \includegraphics[width=\columnwidth]{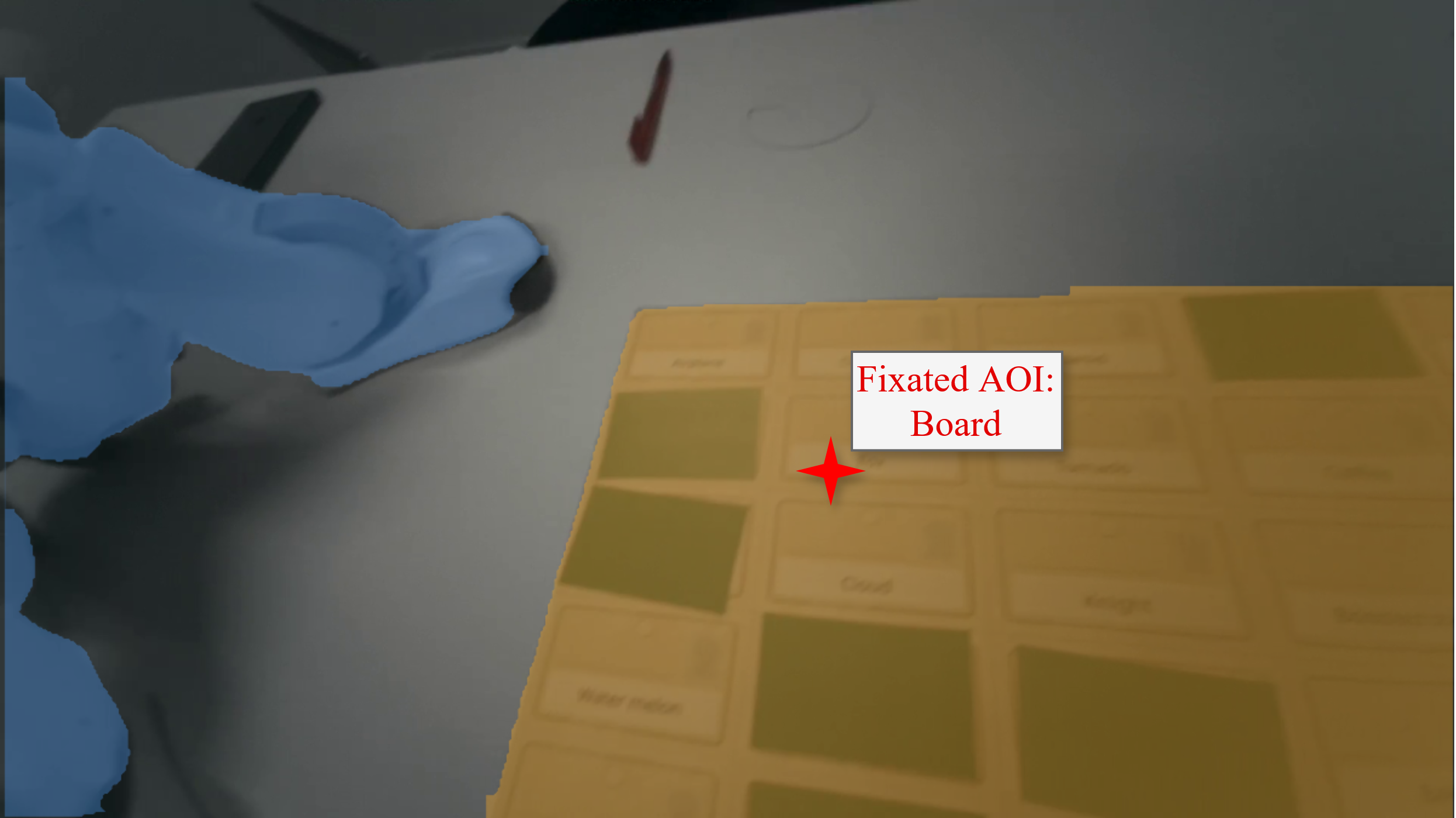}
   \caption{Example frame from the eye tracker's video stream with AOI segmentation masks (blue: robot; orange: board) and the gaze fixation point}
   \label{fig:aoi_annotation}
\end{figure}

\begin{figure*}[ht!]
    \centering
    \includegraphics[width=\columnwidth*2]{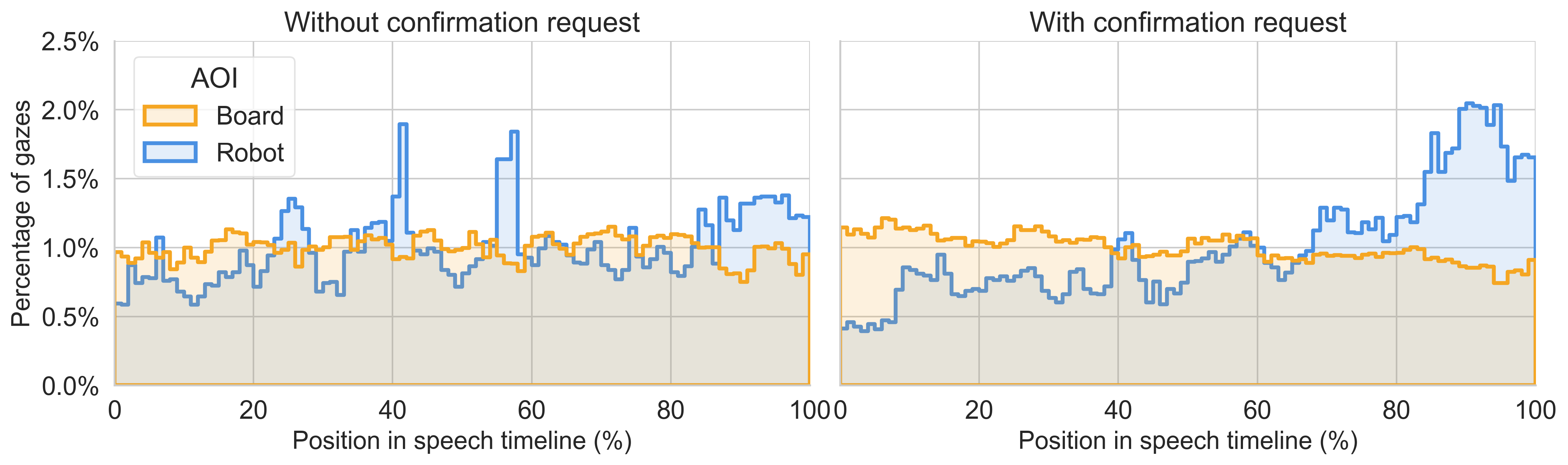}
    \caption{Gaze distribution between board and robot during human speech without (left) and with (right) confirmation request.}
    \label{fig:rq_2}
\end{figure*}

\subsection{Study design and conditions}
The participant's interaction partner is the humanoid robot NAO, which serves as an embodied chatbot with a LLM connection for natural-language responses. The system uses GPT-4o-mini as an LLM with a script that updates the system prompt after each move according to manual input of the game master. For speech recognition, the system uses Faster Whisper as a Speech-to-text model, and Voice Activity Projection \cite{ekstedt2022voice} to mitigate latency in the response behavior of the robot.

The behavior of the NAO robot is manipulated in two conditions: its head direction alternates between the board and the face of the participant during speech. In the mutual-gaze condition (1) the robot predominantly looks at the participant's face, whereas in the referential-gaze condition (2) robot predominantly looks at the board. In both conditions, the robot avoids staring at the participant for longer than 5 s to prevent uncomfortable gaze aversion \cite{binetti2016pupil} and occasionally glances towards the respective other area. All participants experience both conditions in randomized order. The game mechanics were controlled by the game master, including updating the robot's context with the proposed words of the participant using a script. The robot's dialogue behavior was autonomous.


\subsection{Data collection}
We measured gaze data using the wearable eye tracker Tobii Glasses 3 and the accompanied analysis software Tobii Pro Lab. We labeled the gaze behavior by setting custom Times of Interests (TOI) during the human's and the robot's speech, as well as in between the dialogue lines. Speech segments were defined as semantically meaningful utterances, excluding brief backchannels like ``hm'' or ``yeah''. 

In addition to these TOI, we analyzed gaze shifts from robot to human, from human to robot, as well as the human's gaze shift to the proposed word after it has been proposed by the robot. More precisely, we examine gaze behavior when the robot proposes a new word, or when the human asks for confirmation for an own word, of which we set custom events in the timeline of the recordings in Tobii Pro Lab (see Figure \ref{fig:timeline_interaction} for an example). To measure these gaze shifts between robot and board, we defined one Areas of Interest (AOI) for each of them. Using the video stream from the eyetracker's camera and the gaze data output, we ran a custom trained YOLOv8 model to segment these AOIs, obtaining AOI labels for each gaze value. These labels were used to investigate RQ~1. Figure \ref{fig:aoi_annotation} shows an example frame including the AOI annotations. To investigate RQ~2, we measured time to first fixation between the robot proposing a word, and the participant's gazing at this word. More precisely, we measured the duration between the first time a robot mentions a word and the participant's first gaze towards this word.


\subsection{Participants and priming}
We collected gaze data from 10 participants (5 male and 5 female), ranging between 25 and 50 years of age. None of them were native English speakers. Written informed consent was obtained from all participants. We instructed the participants to speak with the robot loud and clear, but otherwise as with a regular conversational partner.

Participants play two rounds, each with a different set of words. We instructed participants to get familiar with the words and their meanings before each round. The clues, their order and the amount of clues are standardized for all participants. The experiment was developed in communication with local ethic authorities. In prior to the experiments, we tested the experimental design twice, each time with a human-human team, to verify feasibility and game difficulty.

\section{Results} \label{sec:results}


To address RQ 1 ``Does confirmation-seeking behavior of the participant trigger increased gaze towards the robot?'', we analyzed the 372 human speech segments in the data, of which 208 contained a confirmation request. Figure \ref{fig:rq_2} shows the distribution of the gaze position AOI across the speech timeline. It's y-axis represents the relative frequency of gazes at each point in the speech, where 1 \% indicates that 1 \% of all gazes occurred at this moment. Participant's gaze distribution towards the robot was slightly more centralized around the end of the speech when they asked for confirmation, than when they did not. In the final 20\% of each speech segment, participants directed their gaze to the robot in 26.7\% of the time when making a confirmation request, in comparison to 20.8\% when they did not. This difference ($M_{\text{diff}} = 5.9\%$) is not statistically significant, $t(9) = 1.817, p = .10$, CI [-1.45\%, 13.30\%], despite having a medium effect size (Cohen's $d = 0.58$).

Regarding RQ 2 ``How does the robot’s gaze behavior affect the time to first fixation towards the target?'', the results indicate that whether the robot is primarily looking at the human or at the board does not influence the time a person first fixates a specific word, with very similar values for both conditions (Mutual gaze: \textit{M} = 1382.9 ms, \textit{SD} = 1246.6 ms; Referential gaze:  \textit{M} = 1377.5 ms, \textit{SD} = 1592.5 ms).

\section{Discussion} \label{sec:discussion}

Our first finding is the difference in gaze distribution ratio between robot and the board: when participants were asking for confirmation, they looked more often at the robot in the last 20\% of their speech than without, though they only looked around 6\% more often to the robot in the mutual-gaze condition than in the referential-gaze condition. Prior art elicited more gaze behavior towards the interlocutor at transition-relevance places \cite{admoni2017social}.
In this game, the board serves as a spatial anchor for the gaze since it contains all the relevant information for the game. Therefore, it is expected that most of the attention is focused around the board. 

We find no effect of robot gaze behavior on the time to first fixation towards a word the robot proposed. This finding aligns with previous works that did not find an effect of reflexive cueing for robot gazes on human responses \cite{admoni2011robot}. However, this finding could also be explained by our task design. The grid of words on the board induces scanning behavior by the participants: when they did not memorize a location of a word, they had to search it first on the board. Although we instructed participants to get familiar with the board before the interaction started, they could still have forgotten the locations of certain words that have been proposed by the robot and that they did not consider before. Another limitation is the manipulation of the robot's head movement behavior, which was perhaps not salient enough to trigger human response behavior.

These results provide initial insights into human gaze behavior in collaborative scenarios where high cognitive load intuitively skews gaze behavior towards task-relevant locations. However, the sample size with \textit{N} = 10 is quite small, reducing the statistical power of the analyses. Additionally, since our experiments did not contain native English speakers, an increased cognitive load during the task may have influenced gaze behavior towards the board.

In future work, we try to expand our experiments to a higher sample size and conduct the experiments in different languages. In addition, we aim to extend the study with equivalent human-human interactions, and study the difference to robot-oriented gazed. Furthermore, we intend to investigate the gaze pattern difference with competitive games and with normal face-to-face dialogue, thus linking our results to the larger body of work on face-to-face dialogues.

\section{Acknowledgment}
We want to thank Lena Hanses for her support during data collection and labeling. Her help was fundamental for the success of this research.





\bibliographystyle{IEEEtran}
\bibliography{IEEEabrv,references}

\end{document}